\documentclass{mva_style}
\usepackage{graphicx}
\usepackage[numbers]{natbib}
\usepackage{amsmath,amssymb}
\usepackage{multirow}       
\finalcopy 

\begin{document}
\title{Panoptic Segmentation of Galactic Structures in LSB Images}

\author{
Felix Richards\\
Swansea Univ,
Dept Computer Science\\
{\tt felixarichards@gmail.com}\\
\and
Adeline Paiement\\
Univ de Toulon,
Aix Marseille Univ, CNRS, LIS\\
{\tt adeline.paiement@univ-tln.fr}\\
\and
Xianghua Xie\\
Swansea Univ,
Dept Computer Science\\
\and
Elisabeth Sola, Pierre-Alain Duc\\
Univ Strasbourg,
CNRS, Obs. astro. Strasbourg\\
}

\maketitle

%

\section*{\centering Abstract}
\textit{
We explore the use of deep learning to localise galactic structures in low surface brightness (LSB) images. LSB imaging reveals many interesting structures, though these are frequently confused with galactic dust contamination, due to a strong local visual similarity. We propose a novel unified approach to multi-class segmentation of galactic structures and of extended amorphous image contaminants. Our panoptic segmentation model combines Mask R-CNN with a contaminant specialised network and utilises an adaptive preprocessing layer to better capture the subtle features of LSB images. Further, a human-in-the-loop training scheme is employed to augment ground truth labels. These different approaches are evaluated in turn, and together greatly improve the detection of both galactic structures and contaminants in LSB images.
}

\section{Introduction}

Recent advancements in astrophysics imaging techniques have made possible the high resolution capture of very faint, or low surface brightness (LSB), galactic structures. Of particular importance are collisional debris (e.g. tidal tails, steams, shells...) which are essential to better understand the history of galaxy evolution. The automatic determination of the physical properties of these objects requires their accurate localisation and segmentation. This task is complicated by the large number of objects (galaxies, foreground stars...) and image contaminants (inc. dust clouds and diffraction figures around bright foreground stars) that are visible in these new LSB images.
Future surveys seek to produce LSB datasets far larger than can be feasibly manually classified, as in \cite{Duc2015TheImages, Bilek2020CensusSurvey, Sola2022CharacterizationImages}. Development of a method to automatically classify LSB structures is therefore crucial for the sphere of LSB galaxy research.

There have been several applications of object detection and segmentation in astronomy.
\cite{Gonzalez2018GalaxyAugmentation,Burke2019DeblendingLearning,Farias2020MaskGalaxies,Levy2021DetectingR-CNN} focused on detecting galaxies. 
Galaxy morphologies were analysed with various levels of detail, from simple classification of morphology types 
\cite{Gonzalez2018GalaxyAugmentation,Farias2020MaskGalaxies}, to classifying the presence of collisional debris \cite{Walmsley2019IdentificationNetworks,Sanchez2023IdentificationNetworks}.

Few works study galaxies in LSB images with deep learning, though their precision tends to suffer from image contamination. In \cite{Sanchez2023IdentificationNetworks} cirrus dust clouds impact the identification of tidal structures, while in \cite{Levy2021DetectingR-CNN} identification of LSB galaxies is confused by tidal structures and cirrus. In \cite{Tanoglidis2021DeepGhostBusters:Images} LSB artefacts such as ghosted halos and scattered light are detected with poor precision. In \cite{Tanoglidis2021DeepShadows:Learning,Walmsley2019IdentificationNetworks} galaxies and tidal structures (respectively) are reliably classified, but cirrus contamination appears weak.
Segmentation of cirrus clouds has been attempted with an ensemble of attention networks \cite{Richards2022MultiSegmentation}, though foreground objects are not a focus in the study.

Building on these previous experiences, we account for image contaminants when segmenting galactic structures in LSB images, through the first combination of the two detection tasks into a unified approach where information is shared. 

Typically, multi-class segmentation is divided into two distinct tasks requiring different methods: instance segmentation deals with foreground objects, suitable for galaxy and tidal structures, whereas semantic segmentation handles background amorphous regions suitable for clouds. \citet{Kirillov2019PanopticSegmentation} propose a novel task, panoptic segmentation, that unifies both segmentation tasks. 
A panoptic segmentation baseline is proposed in \cite{Kirillov2019PanopticNetworks}, where a semantic model, FCN \cite{Long2015FullySegmentation}, is added into Mask R-CNN \cite{He2017MaskR-cnn} so that instance and semantic networks share a common feature generating backbone.
We propose a panoptic model to identify foreground objects and background contamination in LSB images. Following the work of \cite{Kirillov2019PanopticSegmentation} we extend Mask R-CNN with a purpose-designed cirrus segmentation network (Section \ref{subsec:cirrusmaskrcnn}), and further adapt the model to the dynamic range and contrast of LSB images through an adaptive intensity scaling layer (Section \ref{subsec:adaptivelayer}). A training scheme is proposed to mitigate the small availability of LSB data and annotations (Sections \ref{subsec:training1} and \ref{subsec:training2}).

\section{Data}

Our dataset contains 186 MATLAS \cite{Duc2015TheImages} LSB images of average spatial size 6000px$^2$ with two spectral channels. Respectively, 80\% and 20\% of samples are used for training and testing. Each MATLAS image targets a galaxy of interest: we take a 3000px$^2$ crop around the target galaxy of each image and then downsize to 1024px$^2$. Data augmentation is applied as a combination of random flips and 90$^\circ$ rotations, followed by element-wise Gaussian noise with $\sigma=0.1$.

Images were annotated for various galactic structures and image contaminants \cite{Sola2022CharacterizationImages}. We retain five classes: "galaxy", combining main galaxies and companions; "elongated tidal structures"; "diffuse halo"; "ghosted halo", a contaminating diffraction effect from foreground stars; and "cirrus", obstructing dust clouds. Elongated tidal structures include tidal tails, plumes and streams, which appear visually similar and are defined as a propulsion of stellar material from a galaxy. 
We discard image contaminants other than ghosted halos and cirrus, such as satellite trails and instrument artefacts, as they are either very rare or very difficult to correctly predict without a larger field of view. 

Postprocessing of diffuse halo annotations is required to separate overlapping labels. A Euclidean distance transform is applied to the mask to obtain local maximum peaks, representing centres of overlapping shapes. Based on the number of galaxies present, the most separated peaks are chosen. These peaks are used as markers for the watershed algorithm \cite{Vincent1991WatershedsSimulations} to identify the boundary separating each overlapping label. Finally, an ellipse is fit optimally to each separated part.

During the annotation of MATLAS images \cite{Sola2022CharacterizationImages}, astronomers labelled target subsections containing objects of interest - galaxies with tidal structures. Often, in MATLAS images, more galaxies lie outside these subsections which are not annotated. On average, annotators processed 1.7 (std 0.9) galaxies per image.


\section{Method}


\subsection{
Panoptic segmentation model}
\label{subsec:cirrusmaskrcnn}

In this work, we wish to segment cirrus contamination along with localised objects, allowing these two tasks to support each other. Mask R-CNN \cite{He2017MaskR-cnn} is designed to handle the latter category; segmentation of extended amorphous regions such as cirrus contamination fits poorly into this instance segmentation framework
Handling categories of objects that cannot be divided into discrete entities is typically handled by networks of different design, such as FCN \cite{Long2015FullySegmentation} or attention networks \cite{Fu2019DualSegmentation, Richards2022MultiSegmentation}. Thus, there is a strong motivation to extend Mask R-CNN to segment cirrus in parallel to instance segmentation, in a panoptic model.

\begin{figure}[t]
\begin{center}
    \includegraphics[width=0.85\linewidth]{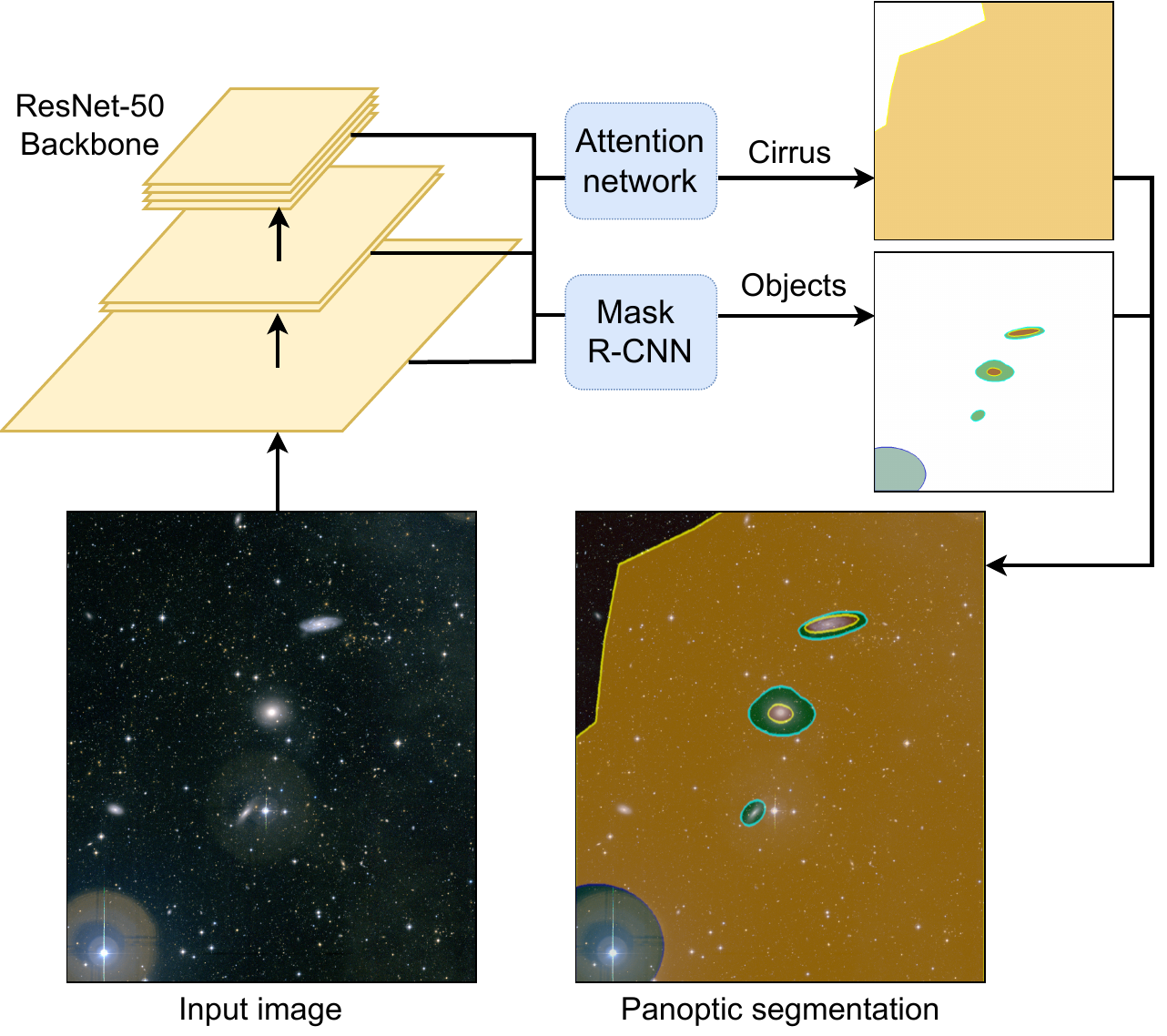}
\end{center}
    \caption{Proposed panoptic segmentation model.} 
    \label{fig:cirrusmaskrcnn}
\end{figure}

We combine the gridded Gabor attention (GGA) network proposed in \cite{Richards2022MultiSegmentation}, designed for cirrus segmentation, with Mask R-CNN, as shown in Fig.~\ref{fig:cirrusmaskrcnn}. GGA performs a computationally efficient multi-scale analysis. It is sensitive to the large scale orientation of textures, which provides a key advantage for recognising cirrus clouds. This is achieved through including a Gabor modulation of features in an attention mechanism. 

The two networks share the same backbone ResNet-50 features, unifying segmentations of discrete objects and of cirrus. Computation along each branch is performed in parallel yielding two segmentations which are overlapped in a multi-label style to achieve a segmentation of all structures in an LSB image.

\textbf{Selection of anchor sizes} - We implement Mask R-CNN with mostly default off-the-shelf parameters (see Torchvision \cite{Marcel2010TorchvisionTorch}). 
While hyperparameter tuning could obtain minor performance improvements, this setup gave consistently good results. Thus, this computationally intensive optimisation is left for future work.

To determine reliable anchor sizes and shapes for our galactic structures, we compute histograms of the heights, widths, and aspect ratios of all target objects. 
The majority of objects have heights and widths between 32px and 512px, and aspect ratios between 0.25 and 2. We therefore set the RPN to consider anchor boxes of widths 32, 64, 128, 256 and 512, and for each box width three aspect ratios are considered: 0.5, 1, and 2. While there are a significant number of object bounding boxes with aspect ratios between 0.25 and 0.5, using unbalanced ratios has the possibility of introducing a bias into the model where tall but narrow objects are favoured by the RPN. This is unwanted as pose variation in localised astronomical objects is naturally balanced (ghosted halos which are artefacts also have unit aspect ratio in most cases). This setup results in 15 total anchor box sizes considered.

\subsection{Adaptive intensity scaling layer}
\label{subsec:adaptivelayer}

A challenge of LSB images comes from their large dynamic range and low contrast of LSB structures. An adaptive intensity scaling layer \cite{Richards2022MultiSegmentation} is placed before the backbone network to exaggerate fainter structures. It implements $\text{arcsinh}$ scaling, popular in astronomy: $X_s=\text{arcsinh}(aX+b)$ with $a$ and $b$ being learnable. We implement 2 layers in parallel, one per image channel, and concatenate their results with the original image to produce a 4-channel input to the ResNet-50 model.

\subsection{Transfer learning}
\label{subsec:training1}

To further improve generalisation of the trained model and reduce overfitting due to the limited sample size, transfer learning is used. Mask R-CNN is pre-trained on MS-COCO \cite{Lin2014MicrosoftContext}. 
While MS-COCO 
is made of natural images 
that appear different from astronomical images, learned features can still be closely applicable on both sets, possibly due to the adaptive intensity scaling layer. Further, this practice is standard in works combining astronomy and segmentation \cite{Burke2019DeblendingLearning, Farias2020MaskGalaxies, Tanoglidis2021DeepGhostBusters:Images, Richards2022MultiSegmentation}. To account for the extra scaling channels, the third convolutional input channel is duplicated. 

For the cirrus 
subnetwork, 
we follow the pretraining of \cite{Richards2022MultiSegmentation} 
on a synthesised cirrus dataset. 
Fine-tuning is performed end-to-end on the complete network.

\subsection{Human-in-the-loop training}
\label{subsec:training2}

A consequence of the non-exhaustive annotation of MATLAS images is that, during training, the model may be penalised for false positives which actually are correct predictions of unannotated objects. To combat this,
we implement a human-in-the-loop (HITL) training protocol 
to iteratively construct a more densely annotated dataset. After an initial training period (30 epochs), we review predictions of all images. Mask predictions of good quality are retained and combined with previous annotations. 
The network is then trained on the new dataset for a shorter period (5 epochs for the following four rounds, then 10 epochs for three rounds to further adapt to a more greatly modified dataset) and new predictions are reviewed, with this process being repeated 7 times. After this reviewing stage, the model is trained 
until a total of 200 epochs is reached, to match the level of learning of a network that would train from scratch on the augmented dataset.

Elongated tidal structures 
were annotated exhaustively in all MATLAS images, therefore they do not require a HITL protocol. Only the galaxy, diffuse halo, and ghosted halo classes are considered.

\section{Results}

\begin{table*}[t]
\caption{Comparison of panoptic and instance detection of objects across different classes from models trained without HITL. Relative improvements of the panoptic model over the instance model are indicated as (+x\%).}
\begin{center}
\begin{tabular}{ll|lllll}
\hline
                          &           & Galaxy & Diffuse halo & Tidal structures & Ghosted halo & All \\ \hline
\multirow{2}{*}{Panoptic} & AP$_{50}$ & 0.782 (+0.5\%)  & 0.788 (+4.5\%)        & 0.000  (+0\%)          & 0.658 (+26.1\%)       & 0.543 (+5.6\%)    \\
                          & AP$_{75}$ & 0.217 (+3.3\%)  & 0.411  (+7.0\%)      & 0.000  (+0\%)          & 0.658  (+34.3\%)     & 0.330  (+21.8\%)  \\ \hline
\multirow{2}{*}{Instance} & AP$_{50}$ & 0.778  & 0.754        & 0.000            & 0.522        & 0.514  \\
                          & AP$_{75}$ & 0.210  & 0.384        & 0.000            & 0.490        & 0.271    \\ \hline
\end{tabular}
\label{tab:panapcomparison}
\end{center}
\end{table*}

\begin{table*}[t]
\caption{AP$_{50}$ scores for standard and HITL models, evaluated on the HITL augmented test data. Relative differences of HITL training over standard training are indicated as (+x\%).}
\begin{center}
\begin{tabular}{l|lllll}
\hline
Training   &  Galaxy & Diffuse halo & Tidal structures & Ghosted halo & All   \\ \hline
Standard          & 0.533  & 0.503  & 0.000 & 0.450  & 0.371 \\
HITL              & 0.797 (+49.5\%) & 0.856 (+70.2\%) & 0.000 (+0.0\%) & 0.814 (+80.9\%) & 0.617 (+66.3\%)\\ \hline
\end{tabular}
\label{tab:hitlcomparison}
\end{center}
\end{table*}

We assess our panoptic model by comparing against separate Mask R-CNN and GGA baselines for the tasks of localised objects and cirrus segmentation respectively on MATLAS LSB images. To ensure a fair comparison, we use the same training protocol without HITL for both the proposed model and separate baselines. 
We also quantify the effect of our HITL training. 

We calculate the average precision (AP) score for various IoU (intersection over union) detection thresholds. 
AP at an IoU threshold $x$ is denoted as AP$_{100x}$, e.g. AP$_{50}$ (threshold 0.5). 
AP is calculated for each class individually and then combined by averaging.

\subsection{Panoptic vs. instance segmentation}

We compare object segmentation performance through a unified panoptic and an isolated instance approach.
To obtain a performance baseline, Mask R-CNN is trained and evaluated on only classes that contain localised objects from the annotated MATLAS dataset. This is similar to the method of \cite{Levy2021DetectingR-CNN}, although they focus on galaxies and do not attempt to detect tidal structures.
Both models are trained for 200 epochs (no HITL is used for this experiment). 

It can be seen in Table~\ref{tab:panapcomparison} that AP scores are increased in the panoptic model for all classes except elongated tidal structures, across all IoU thresholds.

Segmentation of elongated tidal structures proves to be a very difficult task, with neither network making any positive detections of such structures on the test set at any IoU threshold. We verified that the detection succeeds on the training set. Thus, the limited number of training samples and difficult class imbalance is likely the issue. A larger study is required to discern whether or not the panoptic approach improves on the segmentation of elongated tidal structures.

For the galaxy class, the proposed panoptic approach offers a small improvement over only instance segmentation, with AP$_{50}$ increasing by 0.5\% and AP$_{75}$ by 3.3\%. This difference is likely minimal for the easier detection task at IoU threshold of 0.5 as the galaxy core is a strong structure and can be delineated relatively easily even in highly contaminated areas.

The synergistic benefits of tying the tasks of contamination and object prediction is stronger for diffuse halos and ghosted halos, which obtain a more significant increase of AP$_{50}$ scores by 4.5\% and 26.1\%, respectively, and 7.0\% and 34.3\% for AP$_{75}$. This concurs with the previous insight, as boundaries of such structures are impacted by cirrus contamination, and thus the panoptic approach offers a larger synergistic benefit for these objects than for galaxy cores. 

Overall, the benefit of the panoptic approach, quantified through improved AP scores, increases with the IoU threshold, i.e. with increased difficulty for the detection task and required precision of the object delineation. This indicates that predicted boundaries overlap better in the panoptic approach, i.e. correct detections/classifications are of better quality.

\subsection{Panoptic vs. cirrus segmentation}
\label{subsec:res_cirrus}

In addition to improving on instance segmentation performance, the panoptic model also scores higher on the cirrus segmentation task. In \cite{Richards2022MultiSegmentation}, the GGA model scored an IoU of 74.5\% as a standalone predictor on the same benchmark dataset, and 79.0\% as an ensemble predictor where 5 versions of the model were trained and their results averaged to produce the final segmentation mask (see \cite{Richards2022MultiSegmentation}). The same GGA network (without the ensemble prediction), as part of our panoptic model, scored an IoU of 85.5\%, representing a relative increase of 14.9\% and 8.2\% over the standalone and ensemble contamination-only models. Based on this significant increase, it would seem that instance segmentation serves as a significantly beneficial auxiliary task for cirrus contamination segmentation. Given that the two tasks are combined through sharing a feature generating backbone, it follows that the addition of more semantic classes allows the model to generate features that better discriminate between cirrus and non-cirrus pixels.



\subsection{Human-in-the-loop training}

In this experiment, we study how the HITL training protocol further improves the results of our panoptic model. 
In total, 914 objects, out of 1928 apparent false positives, are added into the dataset. Over the first four review stages, 67\% of false positive predictions are added into the dataset. Galaxies, in particular, are reliably predicted, with an acceptance rate of 89\% over these review stages. In the second half of reviewing, the acceptance rate across all classes drops to 41\%, likely as the annotation fields at this stage are saturated with no clear detections left to add.

After training on the extended HITL dataset for a further 120 epochs, we evaluate the model on the HITL-augmented test set. Results are significantly improved compared to the original non-HITL model, indicating that HITL training is beneficial. Indeed, the HITL model evaluated on HITL test data outperforms all panoptic scores by an average of 66.3\%, demonstrating the effectiveness of the HITL training scheme to mitigate against limited data. In particular, performance on diffuse and ghosted halos sustains respectively 70.2\% and 80.9\% higher AP$_{50}$ providing a strong motivation utilising HITL on these classes. 

 

Cirrus segmentation was not significantly impacted by HITL training. 
Thus, while the addition of semantic classes improved the support of features for cirrus segmentation (Section \ref{subsec:res_cirrus}), their refined definition from additional training did not have as large an effect.

\section{Conclusion}

In this paper, a method for automated cataloguing of galactic structures in LSB images was presented. Previous methods often suffer from a lack of unification of localised objects and large image contaminant presented as homogeneous textures. We proposed to address this issue by a panoptic segmentation model where Mask R-CNN was combined with a semantic segmentation network designed for detection of galactic cirrus. The proposed method was used to simultaneously segment galactic structures and cirrus contamination, where we showed that unification of the tasks improved their respective performance. A human-in-the-loop training protocol was utilised to create dense annotations, and significantly improved segmentation accuracy of all detected objects, especially for diffuse and ghosted halos that present more challenging boundaries and overlaps, motivating future studies on active or semi-supervised learning in astronomy.

We demonstrated that fine localisation of galactic structures through segmentation is feasible with deep learning, even in contaminated images. Galaxies and their surrounding diffuse halos, and ghosted halo contaminants were detected reliably, with reasonable delineation boundaries in areas of high cirrus. 
Predictions of elongated tidal structures remains a challenge. Given the good performance on other classes, further investigation is warranted.
We hypothesise that with more training data, automated detection and segmentation of these subtle structures should be possible. 

\section*{Acknowledgement}

This project has received financial support from the CNRS through the MITI interdisciplinary programs.

\balance

\bibliographystyle{plainnat}
\bibliography{bibliography}


\appendix
\section{Training parameters}

The proposed model's instance segmentation branch and the baseline Mask R-CNN model both use SGD with a learning rate of 0.01 which is halved every 25 epochs, and L2-regularisation penalty of $5\times10^{-4}$. The attention-based cirrus sub-network is trained with the Adam optimiser using a learning rate of $10^{-3}$ which is exponentially decayed by a factor of 0.98 per epoch, and L2-regularisation penalty of $5\times10^{-7}$, similarly to \cite{Richards2022MultiSegmentation}. Training time was approximately 12 hours on a single Nvidia GTX 1080 Ti, and inference took $\sim$1s per sample. 

\section{Dataset statistics}

After annotation by astronomers, the dataset contains the following number of labels:
\begin{enumerate}
    \item Galaxy: 256 in total, per image mean=1.64, std=0.91
    \item Elongated tidal structures: 66 in total, per image mean=0.42, std=0.71
    \item Diffuse halo: 305 in total, per image mean=1.96, std=1.18
    \item Ghosted halo: 262 in total, per image mean=1.68, std=1.49
\end{enumerate}
In addition, 48 images out of 186 have cirrus annotation.

After HITL training, the numbers of labels are:
\begin{enumerate}
    \item Galaxy: 439 in total, per image mean=2.81, std=1.64
    \item Elongated tidal structures: 66 in total, per image mean=0.42, std=0.71
    \item Diffuse halo: 529 in total, per image mean=3.39, std=2.06
    \item Ghosted halo: 483 in total, per image mean=3.10, std=2.24
\end{enumerate}

\end{document}